# Application of Computer Vision and Machine Learning for Digitized Herbarium Specimens: A Systematic Literature Review

Burhan Rashid Hussein[a], Owais Ahmed Malik[a,*], Wee-Hong Ong[a], Johan Willem Frederik Slik[b]

[a]Digital Science, Faculty of Science, Universiti Brunei Darussalam, BRUNEI DARUSSALAM
[b]Department of Environmental Life Sciences, Faculty of Science, Universiti Brunei Darussalam, BRUNEI DARUSSALAM

**Abstract**
Herbarium contains treasures of millions of specimens which have been preserved for several years for scientific studies. To speed up more scientific discoveries, a digitization of these specimens is currently on going to facilitate easy access and sharing of its data to a wider scientific community. Online digital repositories such as IDigBio and GBIF have already accumulated millions of specimen images yet to be explored. This presents a perfect time to automate and speed up more novel discoveries using machine learning and computer vision. In this study, a thorough analysis and comparison of more than 50 peer-reviewed studies which focus on application of computer vision and machine learning techniques to digitized herbarium specimen have been examined. The study categorizes different techniques and applications which have been commonly used and it also highlights existing challenges together with their possible solutions. It is our hope that the outcome of this study will serve as a strong foundation for beginners of the relevant field and will also shed more light for both computer science and ecology experts.

**Contents**

---

* Corresponding author.
*E-mail address:* owais.malik@ubd.edu.bn (O.A. Malik).

## 1. Introduction

Herbaria contain plants specimens which have been collected, preserved and documented for future use. They provide the foundation of botanical research and are essential for various studies including plant taxonomy, species variability, extinction risks and phenology [1],[2],[3],[4]. The current estimate of specimens in natural history collections (NHC) is in-between 2-3 billion [5]. Among those 350 million specimens are of dried plants or their parts mounted on a special herbarium sheet [6].

Herbarium specimen have been preserved for several years which is important for regions such as tropics were the amount of these specimen samples are limited [7],[8],[9]. Technological advancement has enabled digitization of these specimens as a way to preserve and make it more accessible to other researchers [10],[11]. These collections become valuable assets especially in tropical regions which are normally characterized with both high diversity and extinction risks [12]. According to Goodwin et al. [13], nearly half of the tropical plants specimens in NHC have been incorrectly identified. Furthermore, an estimate of more than 35000 species have been collected and stored while yet to be described [2]. These are new species with no description either because of lack of necessary expertise, inaccessibility or having incomplete information and hence they have been treated as unidentified. On the other hand, thousands of these specimens have not yet been identified at the species level while others need to be reversed following recent taxonomic knowledge [5]. The threat of species extinction far outnumbers available conservation resources. Lack of enough funds has increased the proportion of species at risk worldwide and the situation set to become worse[14]. Numerous studies have been conducted on possible reintroduction of these species to reduce the risk of disappearance[14],[15]. Traditional manual identification of species involves navigating keys that consist of sequency of prescribed identified steps. In each of these steps, a series of plants characteristics needs to be identified in order to answer prescribed questions. Choosing an appropriate answer may not be trivial [16] and that can lead to inconsistency as this manual process becomes less accurate[17]. More advance methods have been developed such as DNA barcoding but the research has still to gain momentum as most of the reference databases are still immature and the process requires DNA extraction from the harvested tissues which could be destructive [18].

With current existing challenges and the volume of these specimens, it is nearly infeasible for a botanist to execute his/her tasks in a reasonable time. Alternative tools that can utilize herbarium sheet images would be of great help for the botanists and taxonomists working in the herbarium. There has been a growing number of interest in the current digitized herbarium specimens [19], [20], [21], [22], [23], [24]. This is mainly due to the current mobilization to digitize biodiversity data and make it accessible to a larger audience around the world [25].

While there has been a rapid increase in computer vision and machine learning applications for digitized herbarium specimens, this review aims to critically asses the existing literature, current practice, open challenges together with future research direction. The rest of the paper is organized as follows: "Methods" section presents detailed steps used to collect and extract relevant information from the primary identified articles. "Result" sections provides a detailed review of the relevant works. "Discussion" section presents major findings and future research direction.

## 2. Methods

In this study, we followed a systematic literature review (SLR) similar to Wäldchen et al. [26] for analyzing published research articles in the field of computer vision (CV) and machine learning (ML) for herbarium plants specimen images. SLR involves the process of identifying relevant published articles, evaluating, interpreting and aggregating their results to answer particular research questions. In summary, SLR involved three essential steps:

(a) formulating research questions, (b) searching for relevant publications, and (c) using information from relevant publications to answer the research questions [27].

## 2.1. Research Questions

For this study, we tried to answer the following research questions:

**RQ-1:** Demography of publications: What are the venues, geographical coverage of the author's and publication time across primary studies? – The question aims to get a quantitative overview of major frontiers and research groups who are currently active on this topic.

**RQ-2:** Application areas of Computer vision and machine learning: What are the common applications areas of CV and ML techniques for herbarium specimen images? – The aim of this question is to find out what are the major challenges that are been solved by using CV and ML techniques for digitized herbarium specimens?

**RQ-3:** Datasets: What are the currently available herbarium image datasets which can be used in training different machine learning models for different applications? – This question aims to find out how readily the datasets are available for training different machine learning models for different applications highlighted through the second research question.

**RQ-4:** Computer vision and machine learning techniques used: What are the current CV and ML techniques applied to these digitized specimens? – The aim of this question is to analyze current CV and ML techniques which have been applied to digitized herbarium specimens.

**RQ-5:** Prototypical implementation: Is there any software implementation such as a desktop application, mobile app or a web service available to use? – The aims of the question is to analyze how much useful are the current implementations in form of service software platform or mobile application of various CV and ML applications to digitized herbarium specimens.

**RQ-6:** Challenges and opportunities: What are the current challenges and opportunities present in the application of CV and ML techniques for digitized herbarium specimens? – The aim of this question is to analyze the current challenges that are limiting the application of CV and ML techniques and possible research direction that could alleviate these challenges and make these techniques more useful.

Table 1: Seeds paper for forward and backward snowballing

| Study | Publication venue | Topic | Year | ∑ Refs. | ∑ Cits. |
|---|---|---|---|---|---|
| Belhumeur et al. [9] | European Conference on Computer Vision. Springer, Berlin, Heidelberg, 2008 | Roadmap paper for herbarium species identification | 2008 | 25 | 219 |
| Cope et al. [28] | Expert Systems with Applications | Review paper on automated species identification using image processing techniques | 2012 | 112 | 334 |
| Carranza-Rojas et al. [5] | BMC evolutionary biology | Roadmap paper for herbarium species identification using deep learning | 2017 | 41 | 119 |

## 2.2. Data Sources and Selection criteria

We used a snowballing technique to find primary studies related to our topic. This technique involves a combination of backward and forward snowballing strategy. For backward snowballing, given a derived paper from our search strategy, a set of referenced publications in that paper are manually searched as candidate papers for our review process. For forward snowballing, a set of works that have cited the published article are further examined to be included in our review based on our search criteria. The forward snowballing was based on google scholar citations. This mechanism ensures that our search strategy of relevant literature is not confined to certain methodologies, set of journals or conferences or even a certain geographical location. To search for relevant articles, the paper had to comply with our search criteria from the paper title or abstract. This criterion is as follows:

**S1 AND (S2 OR S3 OR S4) AND NOT S5**
Where
**S1:** (herbarium OR herbaria OR "natural history collection" OR "natural history museum" OR specimen)
**S2:** (digitization OR digitisation OR Digital OR digitize OR digitized)
**S3:** ("image processing" OR "computer vision" OR "machine learning" OR "Deep Learning" OR "neural network")
**S4:** (identif* OR recogn* OR classif* OR detect* OR annotat* OR extract* OR segment* OR retriev*)
**S5:** (disease OR insect OR animal OR genetic OR gene OR DNA OR RNA OR protein)

Three studies were identified as a starting point to further searching for relevant papers (Table. 1). To perform the searching process, we primarily used Google Scholar search engine to avoid any possible bias to specific publishers. From the initial results, we then further check if the publication has been listed in one of the following scientific repositories:

- ACM Digital Library
- IEEE Xplore
- ScienceDirect
- Scopus
- Web of Science

This ensured that we only focused on high-quality publications. Using our search criteria resulted in a large number of existing studies. In the next step, we only selected papers that have been listed in any of the mentioned scientific repositories. In the third step, we kept those studies that had some component of CV and ML for digitized herbarium specimens or discussed their potential application. At this point, a total of 59 peer-reviewed primary studies were identified. Finally, we further focused on studies published between 2010 and 2020 to get a more recent overview of the research. This resulted in the selection of 52 primary studies which were then used for performing this review.

### 2.3. Data Extraction

To gather the demographic data for RQ-1, meta-data of the primary studies were used to extract relevant information. We designed a template data extractor that was used to collect the information (Table. 2). Since RQ-2 involved various aspect of the applications of CV and ML techniques, we extracted this information from the primary studies that have already applied CV and ML techniques and other relevant studies that have highlighted the potential application areas. For RQ-3, extraction of relevant information from the primary studies was conducted using the template in Table 2. RQ-4 and RQ-5 involved specific approaches used by the study hence the template in Table 2 was used to extract the required data. To answer RQ-6, the required information was extracted from the primary studies in which authors have highlighted challenges and future solutions to overcome those challenges.

## 2.4. Threat of Validity

In order to ensure the quality of the reviews, it is critical to assess the threat of validity [29]. In this study, the threat of validity arises from two aspects which were the bias in study selection and the bias in the data extraction process. Study selection depends on the search strategy used. To mitigate this bias, we followed the method used in [29] by performing both manual search and automatic search across multiple databases. We used publication title, abstract and keyword to search for our primary studies. However, we limited our search to English-language studies which may have excluded relevant studies in other languages. Nonetheless, the number of studies included indicate the depth of the search. To mitigate the bias in the data extraction process, a data template was created and followed to answer the research questions. Furthermore, two independent extractors were used and whenever disagreement arises other researchers were involved in finding a resolution.

Table 2: Template for data extraction

| | |
|---|---|
| **RQ-1** | |
| Study identifier | |
| Year of publication | [2010 – 2020] |
| Author(s)' country | |
| Author(s)' background | [Biology/Ecology, Computer science/ Engineering] |
| Type of publication | [abstract, conference proceedings, journal] |
| | |
| **RQ-2** | |
| Research applications | Phenological studies, identification, digitization workflows |
| | |
| **RQ-3** | |
| Dataset repositories | |
| No. of species | |
| No. of images | |
| Task | [Optical character recognition (OCR), Identification, object detection, Segmentation] |
| Source(s) | [own dataset, existing dataset] |
| **RQ-4** | |
| Techniques used[a] | [Identification, Segmentation, Optical character recognition (OCR), object detection] |
| (a) Identification | [Plant species, Plant organ] |
| (b) Segmentation | [Whole specimen, plant organ] |
| | |
| **RQ-5** | |
| Prototype name | Name of the prototype if available |
| Type of application | [mobile, web, desktop] |
| Computation | [online, offline] |
| Availability | [public, private] |
| Task | [identification, analysis] |
| | |
| **RQ-6** | |
| Challenges | Current existing challenges identified from the reviewed literature |
| Opportunities | A possible solution to mitigate the identified challenges |

## 3. Results

### 3.1. Demography of publications (RQ-1)

There has been a rapidly increasing interest in using CV and ML for digitized herbarium specimen over the recent years. The number of published research articles in 2020 have doubled over the combined past years and with 35 out of 52 articles involved both computer and non-computer scientists (Fig. 1). Furthermore, 31 out of 52 articles involved both academics (university researchers) and other non-academic researchers (research institutions).

To get more insights into the current frontiers of this research, we analyzed authors by the countries they are representing. Overall, there have been a total of 20 countries represented by these authors with 15 articles written by authors from more than one country. When considering only the first authors country, 24 studies have been performed in Europe, 20 studies have been performed in North America, 6 studies in Asian countries and 2 studies from Africa. Among them, 16 involved non-computer scientist as the first author. Furthermore, from the first authors' county we found out that United State of America (USA) is the leading frontiers with a total of 14 studies followed by German and the United Kingdom (UK) with each having 9 studies, then Costa Rica with 6 studies followed by France having 4 studies and then Brunei with 3 studies, Malaysia and Tunisia each having 2 studies with the last countries of Belgium and Spain each having a single study. It is also worth to note that 36 first authors were computer scientist while the remaining 16 having other academic backgrounds. Finally, we also noticed that all the reviewed journals articles were published in non-computer science or engineering journals.

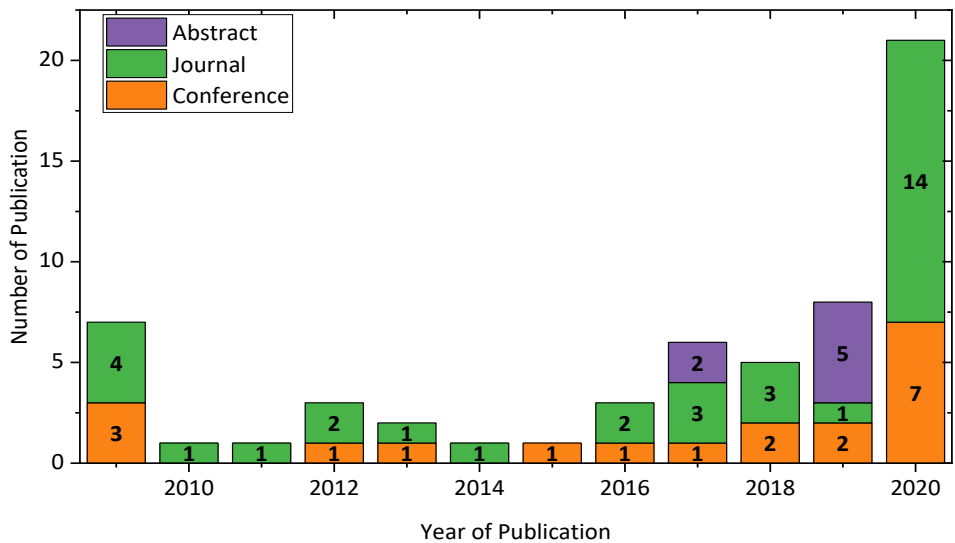

Figure 1: Number of studies per year of publication

### 3.2. Application area of Computer Vision and Machine Learning (RQ-2)

The application of CV and ML for digitized herbarium specimen collections has gradually increased due to massive digital image repositories [30]. From the reviewed literature, we found 23 studies having relatively high focus on phenological features extraction (Fig. 2). These features include leaf measurements [31], morphological traits of specimens [32], detection of plants organs [33] and measurements of reproductive organs [34]. Eighteen studies involved identification of plant species at different taxonomic levels [5], 5 studies have focused on digitization workflow [35] while the remaining studies have focused on studying the interaction between plants and insects [36], reconstruction of damaged herbarium leaves [37], proposing various automation tools, data augmentation technique and the study of reverse senescence for digitization herbarium specimen [38].

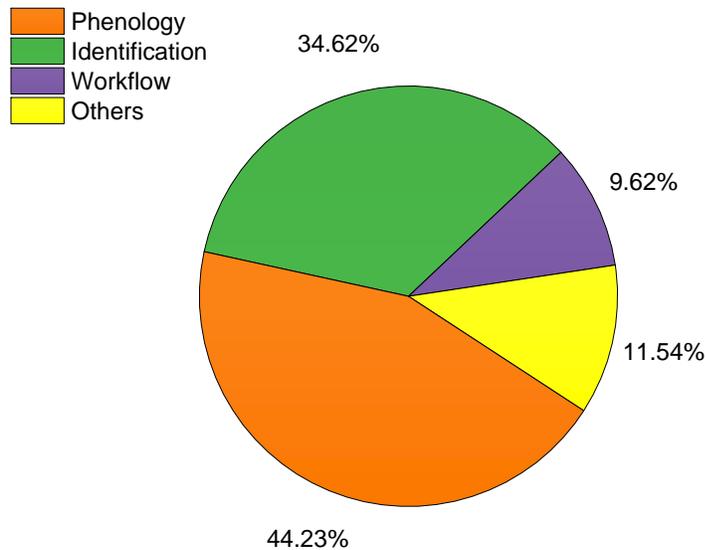

Figure 2: Application areas for computer vision and machine learning for digitized herbarium specimens

### 3.3. Datasets (RQ-3)

Herbaria consist of collections of dried plants specimen with labels that are used as a critical resource for ecological, biodiversity and evolutionary research. For a typical plant to be stored in the herbarium, a fresh sample of a plant is usually collected which undergoes numerous preservation processes including pressing and drying process before storage. These procedures are vital for long time preservation as the presence of water may cause fast damages essentially due fungi, bacteria or insects [39]. Once the plant is dried, it is then mounted on a special herbarium sheet (usually 42 x 29cm) with associated plant information. The process of preservation causes a substantial difference from their field plants as there is a high possibility of color alteration and deformation in the shape as the most of the three-dimensional organs like flower and fruits are flattened out [39]. During specimen digitization, different components such as barcode, color reference charts, scale bar and plant labels together with an envelope containing loose plant material are added to the herbarium sheet. Some of these components are essential to ensure quality images are produced to facilitate other post-processing tasks such as optical character recognition (OCR), natural language processing (NLP) and other image analysis [40], [41]. An example of a complete digitized herbarium plant image can be seen in Fig 3. Details of various components associated with the labels are summarized in Table 3.

Collective initiatives have enabled a mass digitization effort of the current biological collections to facilitate easy sharing and make it more accessible. These efforts are supported by setting up of online digital repositories in which digitized collections can be freely shared. Despite digitization efforts at various institutions, centralized data repositories such as Integrated Digitized Biocollections (IDigBio) is necessary to improve specimen accessibility and diversity across different herbaria. Furthermore, existing digitized collections have rich information that requires manual human efforts to extract relevant features which can be a labor-intensive and time-consuming task.

Table 3: Different component present in a digitized herbarium image

| Herbarium component | Purpose |
| --- | --- |
| Plant label | Plant label consists of bulk information with respect to the species. These information include the scientific name of the specimen, collection location such as longitude and latitude if available, date of collection, habitat information of the specimen and the name of the collector. |
| Color reference chart | Color reference charts are important during the digitization process as they help to ensure the quality of the image produced. |
| Scale bar | A scale bar is normally attached to help provide reference information of the actual physical measurements of the specimen |
| Barcode | It's a unique identifier of the specimen across herbaria which aids in the digitization process. |
| Envelope | The envelope is a folded paper packet that consists of loose plant materials such as plant seeds or fruit which cannot be directly attached on the herbarium sheet. |
| White strips | These are thin pieces of paper that are used to mount and hold the specimen in place. |

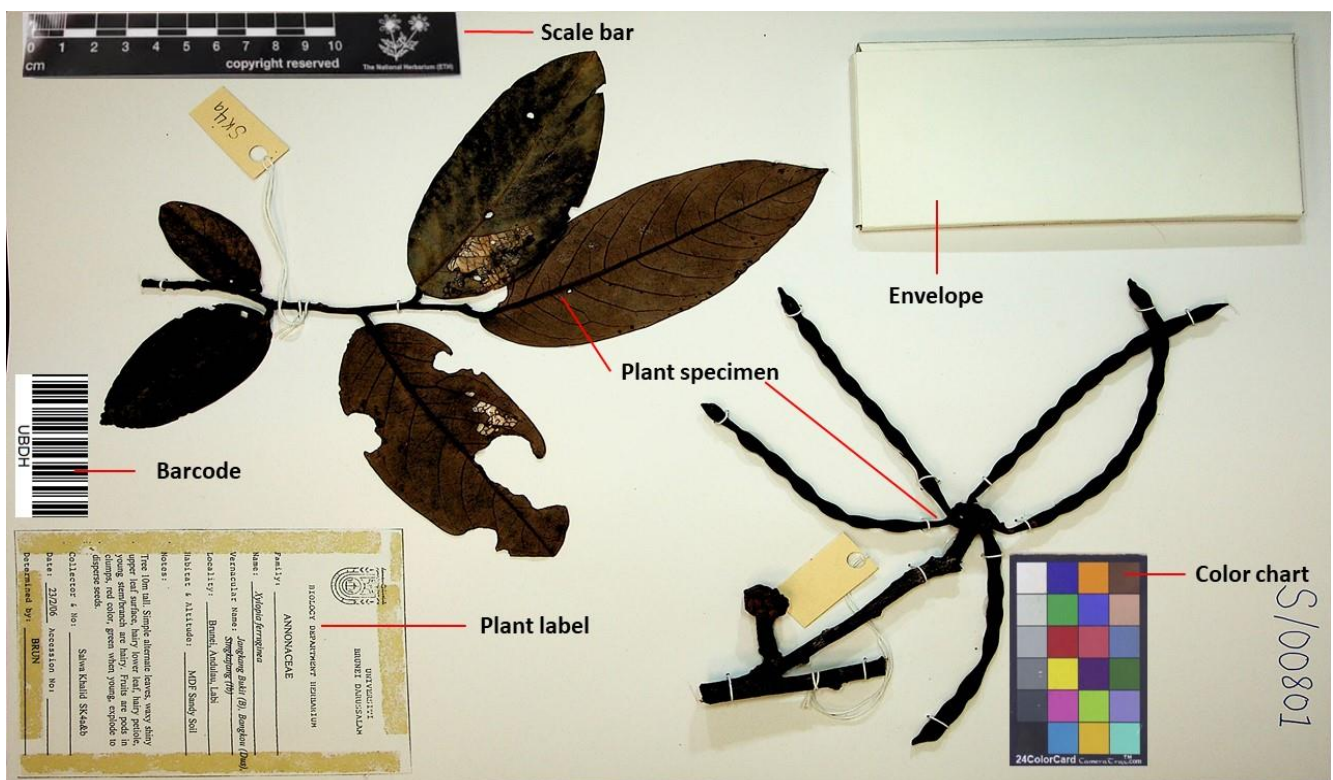

Figure 3: Herbarium sheet

Table 4: Overall image data utilized in the studies

| Task | ∑ Images | No of Taxa | | | Source of data | study |
|---|---|---|---|---|---|---|
| | | Species | Genus | Family | | |
| Identification | 79 | 17 | 1 | - | Private | [42] |
| | 1100 | 4 | 1 | - | Private | [31], [43] |
| | 441 | 4 | 1 | - | Private | [44] |
| | 7597 | - | 2001 | 19 | Private | [45] |
| | 1006 leaves | 95 | - | 3 | Private | [7] |
| | 108 | 6 | - | - | Private | [46] |
| | 441 | 4 | - | - | Private | [47] |
| | 54 | 3 | 1 | - | Private | [48] |
| | 11071 | 255 | - | - | IDigBio | [5] |
| | 253733 | 1204 | | | | |
| | 253733 | 1204 | - | - | IDigBio | [49] |
| | 253733 | 1191 | 498 | 124 | IDigBio | [50] |
| | 46469 | 683 | - | 1 | Public | [51] |
| | 330752 | 997 | - | - | PlantCLEF 2020 | [52], [53] |
| Generative | 120 | 3 | - | - | Private | [38] |
| | 2924 | - | 10 | - | Private[1] | [37] |

Machine learning techniques particularly deep learning offers a new approach to automate these processes. Nevertheless, this approach requires a massive and diverse annotation of data to reach to an acceptable performance [3],[54]. Out of 37 studies, 17 studies have reported results using their own private datasets except for one study which combined both public and the private dataset (Table 4 and 5). Two datasets were curated and hosted online as a benchmarking dataset for herbarium species identification. These datasets include the herbarium challenge 2019 dataset and the PlantCLEF2020 in which herbarium images are used as part of the training data [55], [56].

For PlantCLEF2020 dataset, multiple submission were received although we only found 2 studies that have published their approaches [52], [53] while other approaches were summarized by the competition hosts [56]. This is also the same for the Herbarium challenge 2019 dataset where we found only 1 study that have used the dataset to assess the performance of their proposed architecture [51] while the remaining submissions have also been summarized by the hosts [55], [54]. Out of the 18 studies that have used publicly available datasets, only 3 studies have reported the usage of the dataset extracted from IDigBio, one of the largest online repositories for hosting digitized bio-collection [5], [49], [50]. On the other hand, 2 studies have used the GBIF as a data source

---

[1] https://github.com/Ab-Abdurrahman/ubd-herbarium-repository

[33], [57] another 2 studies have used SERNEC [58], [59] while the remaining 9 studies have made their research data freely available [32], [36], [37], [51], [60], [61], [62], [63], [64].

Among the computer vision tasks, only the identification task has a well-curated dataset in which different authors can use to compare and benchmark their studies [54], [55], [56]. For other tasks such as segmentation and object detection, mostly the authors have only published the original herbarium images and the annotated ground truth have been reported in few studies [33], [65], [58], [61], [63], [64].

Table 5: Overall image data utilized in the studies

| Task | ∑ Images | No of Taxa | | | Source of data | study |
|---|---|---|---|---|---|---|
| | | Species | Genus | Family | | |
| Trait extraction | 103000 | - | - | 11* | Public | [32] |
| | 15554 | - | - | - | Public | [60] |
| | 18389 | - | - | 2 | | |
| | 163,233 | 7782 | 1906 | 236 | Private | [66] |
| | 830408 | 1000 | - | - | GBIF | [57] |
| Object detection | 244 | 2 | - | - | Public | [61] |
| | 653 | 485 | 267 | 59 | GBIF | [33] |
| | 108 | 3 | 1 | - | SERNEC | [58] |
| | 1000 | - | - | - | SERNEC | [59] |
| | 124 | 2 | - | - | Public | [36] |
| | 4000 | - | - | - | Private | [67] |
| Segmentation | 395 | - | - | - | Private[2] | [65] |
| | 2685 | 1000 | - | 165 | Private & SERNEC | [68] |
| | 21 | 1 | - | - | Public | [62] |
| | 3073 | 6 | 2 | - | Public | [63] |
| | 1895 | - | 1 | - | Private | [69] |
| | 50 | - | - | - | Private | [70] |
| | 78 herbarium books | - | - | - | Private | [71] |
| | 400 | 308 | 99 | 30 | Public | [64] |

Among all studies, only 7 studies have reported usage of specimens from different Families [33], [37], [45], [50], [66], [64], [68] while the remaining studies have either not reported or used species from a single Genera [42], [31], [43], [44], [48], [58], [69] except for [63] which reported combined species from 2 Genera. All deep learning approaches for species identification and trait extraction involved relatively large training samples as opposed to segmentation and object detection tasks. This is likely expected as the generation of ground truth annotation is a labour-intense and time-consuming task [72]. More than 7 studies about species identification have used above 5000 images with more than 200 species. For object detection, all studies have used more than 15000 images while extracting various features. For segmentation and object detection tasks, more than 5 studies out of 13 have used greater than 1000 annotations images.

---

[2] https://github.com/Ab-Abdurrahman/ubd-herbarium-repository

### 3.4. Computer Vision and Machine Learning Techniques (RQ-4)

Recent digitization efforts together with advancements in CV and ML techniques have enabled novel research outputs [3]. A typical herbarium sheet consists of a dried specimen (plant) mounted on a preservation sheet accompanied by information labels, barcodes, loose plant parts in an envelope and sometimes a scale bar with a color chart which is used in digitization process [73]. Digitization efforts have motivated increasing use of CV and ML techniques as most of these data are made publicly available. CV and ML generally aim to solve various problem including identification tasks, object detection and segmentation. Figure 4 shows the CV and ML techniques which have been applied to digitized herbarium specimens. Apart from Identification tasks, all remaining CV and ML techniques have used deep learning in solving a specific problem (Segmentation, OCR, object detection and generative tasks).

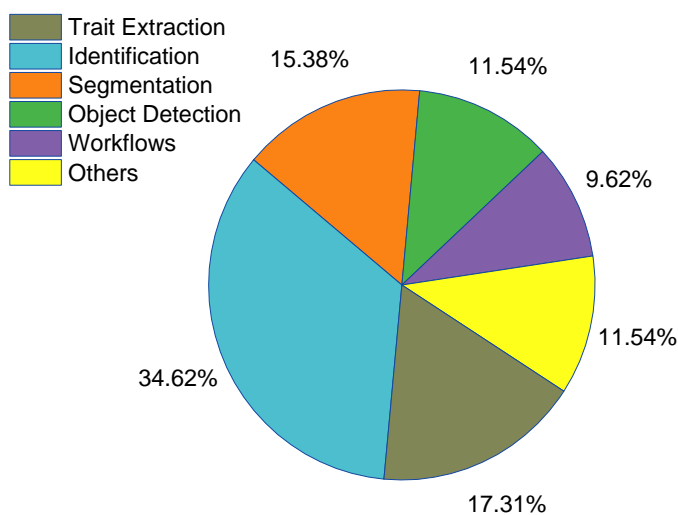

Figure 4: Different computer vision and machine learning applications

### 3.4.1. Identification

For identification task, the model is trained to discriminate between a set of targets. These targets can involve differentiating between plant taxa or other target outputs. Most of the studies have focused at species level identification as this is a more challenging even for botanical experts [54]. Out of 26 studies, 10 studies have used various shape and vein related features to differentiate between plant taxa. Six studies have used deep learning (Convolution Neural Networks) for species identification while the remaining five studies have used the same classification/identification CNN to detect presence or absence of certain features [60], [57], [32].

### 3.4.1.1. Feature Extraction

Feature extraction involves extracting of certain key characteristics of the plant before the identification process. The extracted features are then used to train a classification model to discriminate between target classes. Most of this characters/pattern can be in a form of color, texture or shape [74]. Shape features have been the dominant features as it is used in all 9 studies. Among those, 4 studies have used shape features together with vein and marginal features of the leaves. Features such as leaf vein and leaf margin are usually species-specific hence these are utilized only when available. On the other hand, only 2 studies have combined both shape and texture features. It is no surprise that none of these studies included color features as herbarium leaves are dried plant and hence have lost color characteristics [28], [75].

The shape is considered as one of the most important features for species identification [76]. A good shape descriptor should be invariant to scale, translation and rotation. Studies such as [46] combined scale-invariant shape features (SIFT) and histogram of gradients (HoG) to identify 6 different species of ferns. Using SVM classifier, the study reported accuracy of between 96-100% when a difficult species was referred to an expert to provide an interesting point for extraction with HoG features. Similarly, Wilf et al. [45] used the same SIFT features to identify over 2000 different plant ordinal and 19 different families. The study used herbarium leaves which were chemically bleached to further enhance the vein features. In addition to SIFT features, the authors used a sparse coding technique to compact relevant features and reduce the dimension of the extracted SIFT features. The study reported achieving an accuracy of 73.43% at the family level while achieving an accuracy of 60.4% at the ordinal level. This is the only study that attempted to use a greater number of taxa than any other studies using feature extraction although it did not focus at species level identification due to smaller species samples.

Apart from shape features, only 2 studies attempted to use texture features on herbarium leaves. For example, Unger et al. [77] used a lazy snapping routine tool to extract individual leaves and then applied an image normalization techniques to counteract leaf shape distortion. Key leaf features including leaf convexity, slimness, compactness, rectangularity, perimeter-area ratio, solidity, circularity and the maximum thickness dispersion were extracted. Besides, the study used Weighted HoG to quantify the leaf veins together with Fourier descriptors to extract the texture features. Using an SVM they reported an accuracy of 73% when tested on 24 species and an 84.88% when decreasing the species to 17. Similarly, Jye et al. [48] combined both morphological features such as leaf area, perimeter, eccentricity, varies leaf length, HoG, moments features together with texture features using co-occurrence matrix to train a classifier. The study involved the identification of 3 different species. When using an SVM classifier, the study reported an accuracy of 83.3% which was slightly higher than that of an ANN classifier.

Other remaining studies have combined both leaf margin, vein and other simple and morphological features to perform the identification process. One such study was done by Corney et al. [31] where a set of leaf margin features were used to train an MLP classifier on 4 species from genus Tilia with a total of 1600 images. The features included the total number of teeth, teeth area, tooth frequency, Edge length, the total length of teeth, Length of blade, width of the blade, the perimeter of the blade, are of the blade, compactness, the total number of blade ratio and other derived features from the mentioned features making a total of 22 features. This study reported a performance accuracy of only 44%. Based on this study [31], Corney et al. [44] combined additional morphological features such as length, width, perimeter, lobe length, area, number of teeth, teeth area, tooth angle, tooth frequency, edge length and other derived features to train an MLP classifier. Using a total of 41 features extracted from 441 leaves, the accuracy improved to 56%. Due to lower performance obtained despite increasing the number of features, Clark et al. [47], used both MLP and a self-organizing maps (SOM) to remove noise data and eliminate some features which were found to inhibit the performance of SOM. Using a total of 441 images from 4 species with only 21 features, the author reported an improved accuracy to 95% when using MLP.

The use of HoG in extracting leaf vein features did not prove to be sufficient to identify species in [77]. But, Hussein et al. [7] showed a significant performance improvement when using only vein features. The authors manually extracted a set of 15 features including different vein angles, leaf width, length and distance between veins. Using an LDA classifier, the reported performance was 85% accuracy for family Dipterocarpaceae which had a total of 63 species. However, the performance drops to 79% when the features were used to identify 15 species from family Euphorbiaceae. When the same features were used to identify 17 species from family Annonaceae, the perform was only 56% accuracy. The work suggested that the vein features are strong indicators to differentiate among species in family Dipterocarpaceae which is not the case for species of family Annonaceae. Automatic extraction of the vein feature could further enhance the usability of these features in other plant taxa. In another study, Wijesingha et al. [42] used only leaf width, length, perimeter and leaf area to identify between 30 species from family Dipterocarpaceae. The study used a probabilistic neural network (PNN) and achieved an accuracy of 85%.

Despite the reported performances on the reviewed studies, most of the used shape descriptors are too simplistic for a large number of species as they are not invariant to scale. On the other hand, It is likely that the leaves used have been collected at different time, maturity stages and maybe sampled from the same herbarium sheet which may create biases in the reported results [78]. With exception of [46], all remaining studies have been applied on individual leaves which were manually extracted from herbarium sheet. Automation in leaf extraction could further boost the number of samples involved and hence provide a more realistic performance of the feature used.

### 3.4.1.2. Deep Learning Approaches

An image is considered as a set of raw pixels with different intensities that represent an entity/object in the real world. Since most of the images have a 2D representation, various CV and ML techniques have been developed to extract meaningful features for a particular task which is popularly known as hand-crafted feature engineering. Despite the success of handcrafted features, the emerging of deep learning approaches specifically convolutional neural networks (CNNs) have shown exemplary performance in various computer vision tasks [79], [80], [81], [82].

CNN can model the spatial or temporal relationship that exists within the data. A typical CNN consists of convolution layers with a certain number of filters/kernels, non-linear processing units to introduce non-linearity in a feature space together with sub-sampling layers for reducing the dimension of features and making the features invariant to various geometric distortions. Figure 6 shows an example of a simple CNN used for image classification. Notice that a fully connected layer is sometimes added at the end of last convolutional or sub-sampling layer to enhance the non-linear combination of all extracted features from the previous layers. Besides, other layers such as batch normalization and dropout layers are added to enhance the performance of the network. The fundamental performance of CNN architectures comes from a novel design of different CNN components. The key features of a CNN are its ability to automatically extract meaningful features, hierarchical learning, ability to share weights and multitask learning capabilities.

We identified 11 articles which have used CNN to perform different plants identification tasks. Six studies have used CNN models to identify herbarium species, 4 studies have used CNN to identify/extract different specimen trait and 1 study which performed both species identification and trait extraction. Out of all 11 studies, only 1 study [60] has attempted to develop their own CNN architecture from scratch while the remaining studies have used pre-existing models.

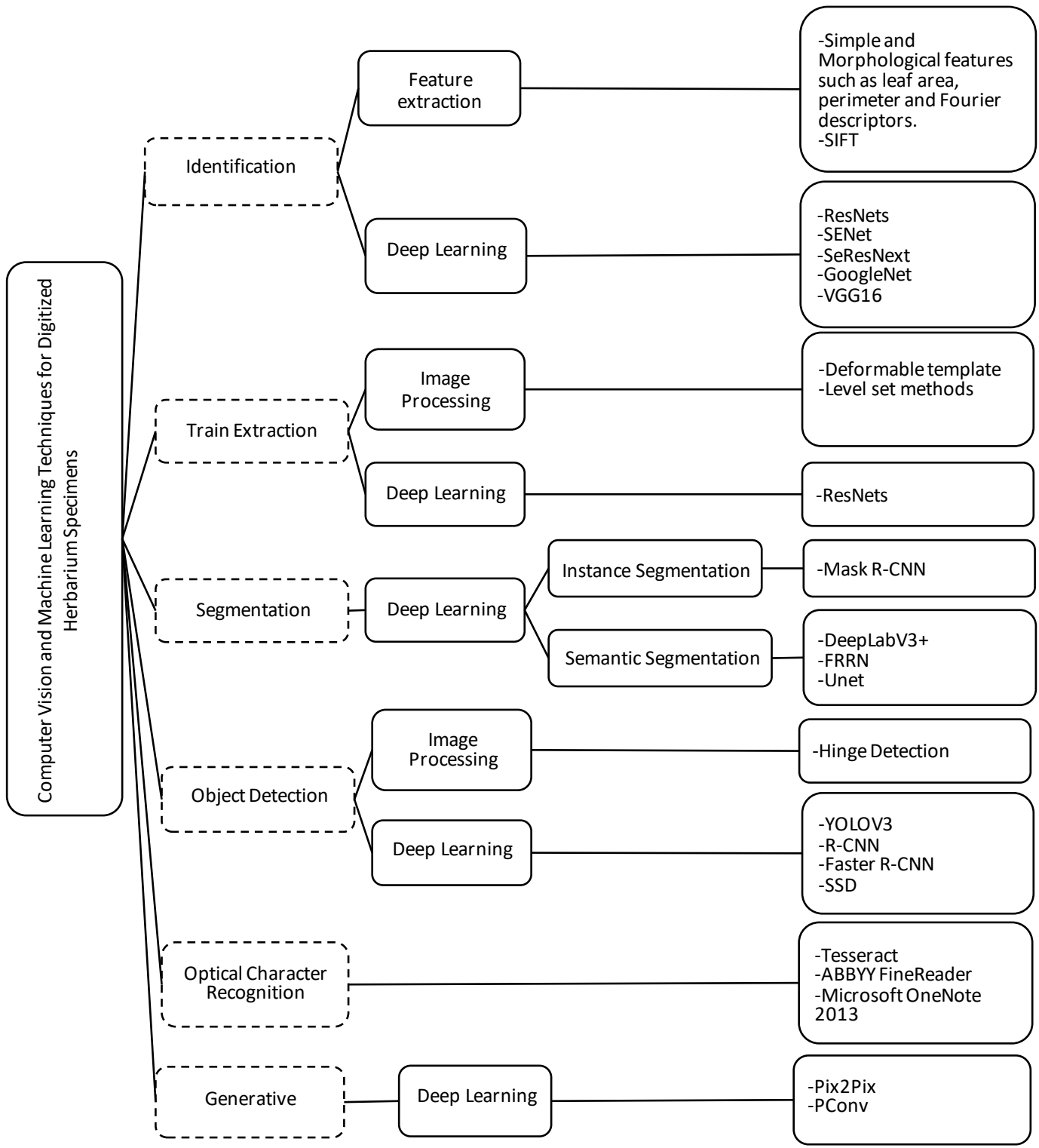

Figure 5: Categories of different computer vision tasks (dashed box) and common techniques (solid box) found in the literature

**3.4.1.2.1. Species identification**

One of the earlier works to use deep learning for species identification was performed by Jose et al. [5]. In their study authors explored the possibility of transfer learning from two aspects: (1) possibility of transfer learning between herbarium collection from different locations and (2) transfer learning between herbarium collection and field plants. The study involved a total of four different datasets including two herbarium sheet datasets with 1204 species and 255 species respectively from different locations. Another two datasets were of fresh plants; one with 255 species of leaf scans and the other with 1000 species from PlantCLEF2015 dataset [83]. Using a modified GoogleNet architecture the network was trained with a cropped portion of 224x224 herbarium image. The study reported several important findings: (1) transfer learning between herbarium improved the model performance although using ImageNet weights with herbarium weights yielded a better performance (2) the transfer learning between herbarium to field images was counterproductive as the accuracy decreased even for plants from the same species and (3) transfer learning across herbarium even with different species proved to be beneficial which could facilitate the application of deep learning techniques in areas with under-represented species. The study reported an overall performance of 70.3% and 79.6% accuracy for herbarium dataset with 255 and 1204 species respectively. The performance of field images was low with 52.3% accuracy for 1000 species and 51% accuracy for leaf-scan with 255 species.

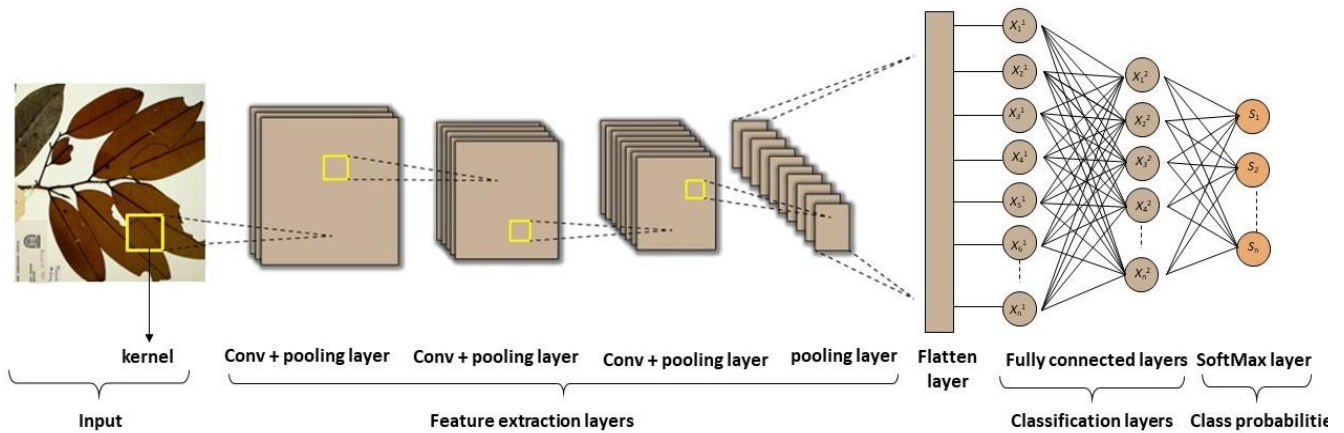

Figure 6: Convolutional neural network for image classification

Jose et al. [50] further explored the use of hierarchical identification process by incorporating higher taxonomic information. The study explored three different architectural setups, (1) the baseline model for an individual family, genus and species identification, (2) a multi-task model where the same model is trained with three different classification heads for family, genus and species (3) and the third model was the hierarchical output of the previous taxon, used in predicting the output. The study used a total of 253,733 images from 124 families, 498 genera and 1191 species. Using an input image of 224x244 with a modified GoogleNet architecture, both multi-task and hierarchical models outperformed the baseline model by a larger margin. The study suggested that weight sharing strategy could further improve the performance in species identification.

Availability of benchmarking herbarium species identification datasets have started to show various performance improvements over the previous deep learning models. For example, the dataset curated by Tan et al. [55] received the best performance of 89.8% accuracy by the winning team. This dataset consists of a total of 46,469 herbarium images from 683 species of Melastomataceae family. The best submission trained ensemble of different deep learning models including SeResNext-50, SeResNext-101 and ResNet-152. All models were trained on a cropped 448x448 image using standard augmentation techniques with various loss including cross-entropy, focal loss and class-balanced focal loss. All models were pre-trained with both ImageNet weights and the INaturalist challenge dataset which consists of field plants. Similarly, Touvron et al. [51] proposed a train-test strategy to improve the performance of the classifier. The authors suggested to train a classifier with smaller resolution images and apply a simple fine-tuning process for the test image resolution. The proposed method was tested on the Herbarium challenge dataset [55] where the authors trained two ensemble models(SENet-154 and ResNet-50) with a training resolution of 448x448 and 384x384, respectively. The network was then tested on a 707x707 and 640x640 resolutions in which the ensembled achieved an accuracy of 88.845% with averaged results from 10 image crops.

### 3.4.1.2.2. Cross-domain species identification

Massive digitized species of herbarium specimens are now being appreciated for a different purpose. For example, a new benchmarking dataset for species identification is the PlantCLEF 2020 dataset which proposed to combine both herbarium and field images [56]. The datasets aimed to introduce a cross-domain species identification between field plants and herbarium plants where herbarium images can be used to train an identification system to identify species from field images. Villacis et al. [53] proposed a domain adaption technique using techniques called Few-Shot Adversarial Domain Adaption which aimed to create a new feature extractor space that is independent of both domains. The study used various augmentation methods including rotations, color jittering, flipping, centre cropping and centre tilting. Using a ResNet-50 encoder with a custom discriminator and CNN classifier, the authors reported performance of only 0.18 and 0.052 mean reciprocal rank (MRR) for the two separate test sets. Similarly, Chulif et al. [52] proposed a Convolutional Siamese Network to learn feature similarity that exists between herbarium images and field plants. The study used a modified InceptionV4 architecture with triplet loss and various augmentation techniques such as flipping, color distortions (brightness, saturation, contrast) and cropping. Using a 299x299 image size, the study reported performance of 0.121 and 0.111 MMR for the two test sets. Compared to the approach of Villacis et al. [53], the proposed method by Chulif et al. [52] generalized better for the difficult test set.

### 3.4.2. Trait Extraction

Since the 19th century, herbaria have preserved millions of specimens which are the only tangible evidence to study plant phenology [62]. Out of eight studies, three studies proposed their approaches using different image processing techniques while the remaining five studies have used CNN to perform various trait extraction from digitized herbarium specimens.

### 3.4.2.1. Image Processing Approaches

Corney et al. [69] demonstrated a leaf character extraction pipeline by using various image processing techniques. The pipeline works by first applying segmentation algorithms including binary thresholding, canny edge detector and active contour model. In the next step, the leaves are identified using deformable template approach and candidate intact leaves are selected by comparing centroid contour distance with the ground truth hand-labelled leaves. Using the proposed pipeline, the author extracted a total of 1645 leaves using 1127 herbarium images from genus Tilia. The authors then extracted various morphological features including leaf length, width and area. As a continuation of their work, Corney et al. [31] proposed a new algorithm to extract leaf marginal features. The algorithm was able to automatically locate leaf teeth and extract features such as tooth perimeter, area, internal angles as well as counting them. Centroid distance was used to locate teeth on the margin of the leaf by successful measuring the difference between the leaf centroid and the point on the leaf margin. Once located each tooth was represented as a triangle and other required features were then extracted. The authors extracted a total of 41 features including other morphological features such as leaf area, perimeter, length, and other derived features [44].

### 3.4.2.2. Deep Learning Approaches

Deep learning techniques have recently received more attention for trait analysis of digitized herbarium specimens [10], [84]. Schuettpelz et al. [60] proposed CNN models to identify two different plant families and detect whether the specimen was mercury stained or not. The study used a total of 18,389 images to train custom CNN models with a resized image of 256x256. The study reported an accuracy of 90% for detecting mercury-stained images while a 96% classification accuracy for plant families. Similarly, Lorieul et al. [66] proposed a CNN model to detect the presence of fertile material from digitized herbarium specimen. The specimen was considered as fertile if the model detects the presence of reproductive structure such as flowers, sporangia (ferns), fruits(angiosperms) or cones (gymnosperms) [66]. The study trained a modified ResNet-50 model with a total of 163,233 specimen images. Using input resolution of 400x250 with random rotation and flipping, the authors reported an accuracy of 96.3%. The study further used a subset of 20,371 images to perform a fine-grained flower and fruit detection. The study reported classification performance of 84.3% by training the same model with an input image of 850x550.

Younis et al. [57] performed both species identification and detection of various leaf traits from digitized herbarium specimen. The study involved more than 1000 species with a total of 830,408 images. By using a centre crop with a 225x225 resolution, the study trained a modified ResNet model [85] and achieved a performance of 82.4% accuracy. On the other hand, the authors further trained modified ResNet model with a sigmoid head to detect 19 various leaf traits such as types of leaf arrangement, leaf form, leaf margin, leaf structure and leaf venation. Using the same preprocessing strategy by centre cropping and resizing an image to 225x225 the model achieved an accuracy of 89.6%. Similar to Younis, Zhu et al. proposed another CNN model to extract morphological traits from digitized herbarium images [32]. The authors used a modified ResNet model to train the network in detecting 4 different leaf traits including leaf margin, leaf arrangement, leaf attachment and if the plant was woody or not. The model reported an accuracy of 80% using the training data collected from 11 different taxa with a total of 103,000 images.

### 3.4.3. Segmentation

Image segmentation has been a long-term computer vision problem and has been attempted with different algorithms such as image thresholding, watershed algorithms, graph partitioning methods, K-means clustering and many others [86]. CNNs in image segmentation tasks have received much attention due to their good performance in image classification tasks [87]. Segmentation is more challenging as it involves both object detection and localization which is achieved by assigning labels to every pixel in an image. The segmentation task can further be divided into three categories which include semantic segmentation, instance segmentation and panoptic segmentation. Both semantic and instance segmentation processes involve pixel-level annotation, although difference comes while assigning label on multiple objects of the same class. For example, in semantic segmentation, multiple objects belonging to the same class are treated as a single entity while in instance segmentation the network is trained to treat multiple objects of the same class as distinct objects which makes instance segmentation more challenging. On the other hand, panoptic segmentation encompasses both semantic and instance segmentation where the network is trained to assign pixel class labels and segment all object instances uniquely [88]. In this review, most of the studies have focused on applying deep learning techniques for segmentation tasks. Specifically, these studies have proposed and used both semantic and instance segmentation for digitized herbarium specimen.

We identified 12 primary studies which have used deep learning segmentation techniques to automate various segmentation tasks including phenological feature extraction and automatic digitization workflow. Among studies, 9 have used semantic segmentation techniques while the remaining studies have used instance segmentation. In the next section, a brief overview of the segmentation technique is given focusing on the current CNN architecture that has commonly been adopted in these studies.

### 3.4.3.1. Semantic Segmentation

Semantic segmentation has been widely adapted with either a new domain area of application or improvements in existing architectures [89],[90]. The architecture involves two main parts. The first part is the encoder network which uses a modified CNN without the full connected layers to develop a low-resolution feature map of the input with higher efficiency in discriminating between classes. The second part (decoder network) upsamples the learned feature map into a full-resolution segmentation map to provide a pixel-level classification. Figure 7 shows an example of an encoder-decoder architecture.

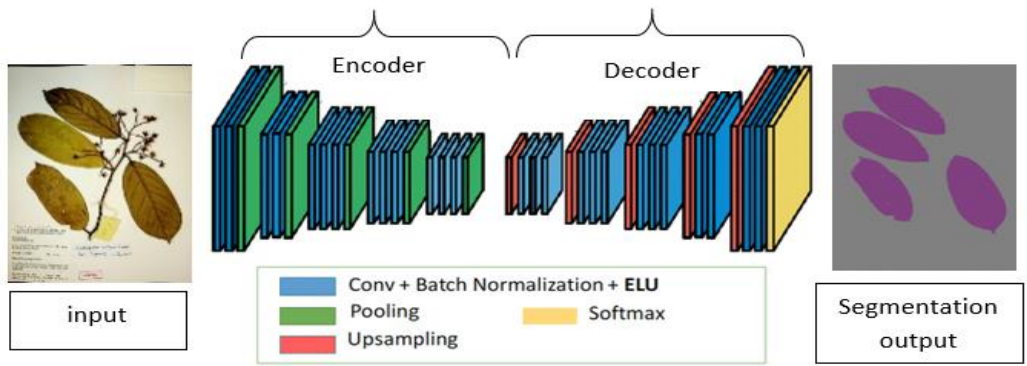

Figure 7: Fully convolutional network for semantic segmentation

From the reviewed literature, DeepLabV3+ has been found a popular semantic segmentation model for plants' leaves segmentation [65], [68]. Deeplabv3+ follows the same encoder-decoder architecture as shown in Fig 7. In the encoder phase, deeplabV3+ uses pre-trained CNNs which have been trained for image classification tasks such as ResNet or VGG16. DeepLab families use Spatial Pyramid Pooling to process input image at multiple scales to capture multi-scale features and later fuse the output to produce a feature map [91].To improve its efficiency, an Atrous convolution operation was introduced. This operation enables the window size of the kernel to expand without increasing the number of parameters [92]. This expansion of window is controlled by the dilation rate and it enables the network to capture information from a larger receptive field of view with the same parameters and computational complexity as the normal convolution. The combination of spatial pyramid pooling together with Atrous convolutions resulted in an efficient multi-scale processing module called Atrous Spatial Pyramid Pooling (ASPP).

In the earlier version (DeepLabV3) [93], the last ResNet block of the modified ResNet-101 uses different Atrous convolution with different dilation rates. ASPP together with bilinear upsampling is also used on top of the modified ResNet block. DeepLabv3+ is an improvement of the previous version by enhancing the decoder module to improve the boundaries of the segmentation [94]. Furthermore, apart from ResNet-101, Xception model can be used as a feature extractor while using a depthwise separable convolution to both the decoder module and the ASPP hence improving the robustness and speed of the network. Intersection Over Union (IoU) has been the most common metric for evaluating semantic segmentation models as it quantifies the percentage overlap between the ground truth mask and the predicted mask [80]. The metric calculates the intersection between the predicted and the target pixels over the sum of their union.

Hussein et al. [65] proposed a semantic segmentation model to remove existing visual noise as a preprocessing step before further processing the specimen. The study compared two popular deep segmentation model with one being pre-trained DeepLabV3+ with pre-trained ResNet-101 as the network backbone against the Full-Resolution-Residual-Network (FRNN-A) which was trained from scratch [95]. The authors manually annotated a total of 395 herbarium images with two classes (specimen, background) for training the models. FRNN-A achieved an IoU of 99.2% while DeepLabV3+ achieved a slightly lower IoU of 98.1%. Despite FRNN-A outperforming DeepLabV3+, the authors recommend DeepLabV3+ as it was more efficient in terms of computational complexity with a negligible sacrifice in accuracy. Similarly, Weaver et al. used DeepLabV3+ with ResNet-18 as a backbone to automatically segment leaves from the rest of objects on an herbarium specimen image. The authors manually annotated a total of 425 images randomly selected from different herbaria with five different classes (leaf, stem, fruit/flower, text, background). The overall performance of the model was IoU of 88.8% for all classes while for leaf classes dropped to only 55.2%. The study further used a trained SVM classifier to identify individual intact leaves from the segmented leaves.

Similar to the work of Hussein et al, White et al. [64] proposed a U-Net model to automatically generate segmentation masks from digitized herbarium specimens of fern taxa. The study used a total of 400 images and achieved a Dice's coefficient of 0.96.

While other studies have focused on specimen segmentation, Hidalga et al. [70] proposed deep semantic segmentation model to automate both image quality management (IQM) and information extraction. The authors hypothesize that the use of segmentation model can speed up the process of IQM and information extraction as these processes will be applied on a portion of the image (segmented parts) rather than the whole image. The authors used an existing model[3] to test their proposed approach on a total of 250 test images for information extraction with OCR and 50 images for IQM. The study reported a speed-up of 49% when using the segmented outputs of the model compared to processing full image. On the other hand, the process of IQM took a total of 34 minutes from the 56 segmented barcodes and 50 color charts compared to 100 minutes which the process took when processing whole images.

### 3.4.3.2. Instance Segmentation

Instance segmentation is considered as a more challenging computer vision problems as it involves the task of both detecting and precisely describing each distinct object of interest that appears in an image. Despite the challenges, instance segmentation remains to be an important computer vision task due to its potential of applicability when it comes to object occlusion/overlapping [89]. With the advancement of deep learning specifically CNNs, numerous instance segmentation framework has been proposed [80], [96]. One popular framework is the mask region-convolutional neural network (Mask R-CNN) [97]. The Mask R-CNN framework offers a straightforward and efficient approach for both detection and instance-level segmentation (Fig. 8). The framework evolved from its predecessors for object detection (Fast/Faster R-CNN) [98]. A fully convolutional network (FCN) was then added along the box-regression and classification to predict objects mask. To improve the performance of the framework, a backbone feature extractor such as feature pyramid network (FPN) has been commonly used as it provides a strong combination of both high- and low-resolution semantic features in a top-down manner with lateral connections [99]. Like semantic segmentation, instance segmentation is commonly evaluated using mean average precision (mAP) metric [96]. The mAP metric measures the overlap between the predicted mask and the ground truth mask at different confidence interval usually between AP.50 to AP.95.

---

[3] https://github.com/NaturalHistoryMuseum/semantic-segmentation

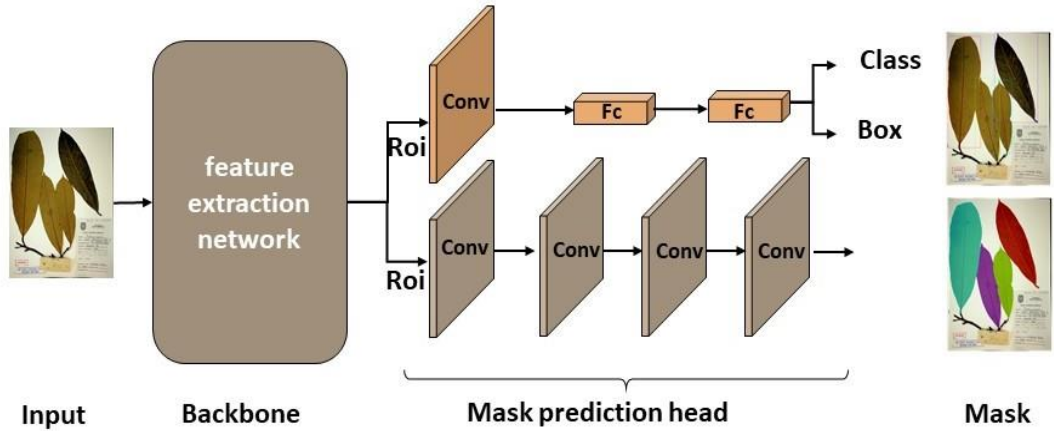

Figure 8: Mask R-CNN framework for instance segmentation

Out of 12 studies which have applied deep learning segmentation models, only 3 studies focused on using instance segmentation. This is likely because of the labour-intensive task required in the annotation of the training data. Mora-Fallas et al. [34] proposed an integration framework to include an active learning mechanism with Mask-RCNN model. The proposed system enables alternation between manual and automatic annotation based on the training process. Further, the active learning process can be tuned to estimate the kind of objects which need manual annotation in the next learning cycle. Using the proposed framework, the authors annotated more than 10,000 reproductive organs including flower, buds, fruit and immature fruits of a species Streptanthus tortuosus which is known for have varying appearance and hence difficult to process with deep instance segmentation model. Similarly, Goëau et al. [62] investigated on three different types of annotation for training Mask-RCNN model including (i) a simple point on an organ (ii) a partial mask which does not cover the whole organ and (iii) a full mask which covers the whole organ of the plant. The study involved four different reproductive organs including buds, flower, fruit and immature fruits of a species Streptanthus tortuosus. Using a total of 21 herbarium images with over 1036 manually annotated reproductive structure with ResNet-50 backbone and a Feature Pyramid Network, the study reported an accurate estimate of 77.9% for all reproductive structure however, the model was more successful in segmenting larger reproductive structure such as flower compared to smaller one like immature fruits. On the other hand, the use of full mask for training the model was more beneficial although the authors urge that the use of point mask can be useful in producing partial mask which can improve the performance of the model. On the other hand, Davis at al. [63] proposed using Mask-RCNN model for counting reproductive structure in digitized herbarium specimen. The study focused on three reproductive structures including fruits, flowers and buds from six common wildflower species found in eastern United State using more than 3000 specimens' images. The trained model achieved an overall accuracy of more than 87% on all reproductive structures and it was more than 90% accurate while estimating an individual reproductive structure using ResNet-50 as a backbone with Feature Pyramid Network. Despite that the deep learning model have been performing better than crow-sourced data, these models are unable to surpass the expertise of Botanical Experts.

### 3.4.4. Object Detection

Automating information extraction from digitized herbarium images is often difficult due to the existence of visual noise. Techniques such as semantic segmentation or instance segmentation offer a more fine-grained localization of the required features but present a challenge of their own as they require a pixel-level annotation of training data which is a labour-intense task to produce. On the other hand, object detection techniques offer a coarse localization of the required features using a simple bounding box annotation of training data which is typically faster to produce than a pixel-wise annotation [98], [100]. A naïve approach to object detection involves the classification of an object within a certain window size of an image. This approach is computationally complex as it requires precise window size for multiple objects on the same image which is hard to attain. Deep learning techniques overcome the aforementioned drawback as the network is trained to localize multiple objects with varying size and location within the same image. Among deep learning models for object detection, Single Shot Multibox Detector (SSD) [101] has been used by [36], [58]. On the other hand, Faster region-convolution network (Faster R-CNN) model [102] has been used in [33], [59] and [61] while Triki et al. [67] have used the recent state-of-the-art YOLOv3 model [103]. Like instance segmentation, mAP is a common metric for evaluating object detection models.

In contrast to deep learning technique, Chandrasekar et al. [71] have developed a custom algorithm for object detection using traditional image processing methods. Since old herbarium specimen were stored in a book line storage, the authors proposed a custom algorithm to detect and correct deformation that may arise during the digitization process. The algorithm works by detecting the herbarium page book using various image processing techniques and then performs hinge detection to ensure only one page is extracted at a time. Finally, the algorithm performs morphological corrections to correct page deformation and makes the final page as flatten as possible. Using their proposed algorithm, the authors demonstrated a good generalization of the OCR system against an original herbarium book without page detection and correction.

Meineke et al. [36] proposed a deep learning object detection technique based on Single Shot Multibox Detector (SSD) to detect and classify 6 different herbivory insect damages from digitized herbarium specimens. The authors manually annotated a set of 124 herbarium images belonging to 2 plant species of Quercus bicolor and Onoclea sensibilis using a VGG image annotation tool to generate a total of 6616 bounding boxes. All annotated images were resized to a 224x224 and were split into training, validation and testing. The study first trained the SSD network to detect two categories of insect damages. To enhance the performance of SSD network, the authors generated hard negative examples (undamaged leaf parts) to train along the annotated b-boxes. From the detected b-boxes, a simple classifier based on pre-trained VGG16 network was trained to further classify the type of damages. The study reported an overall MAP of 45% for the SSD while achieving an accuracy of 81.5% for the classification network.

Similarly, Pryer et al. [58] proposed an object detection and classification pipeline that was trained to discriminate three species closely related to genus Equisetum (E. hyemale, E. xferrissii and E. laevigatum). From the pipeline, the authors trained an SSD object detection model with only two species data of E. laevigatum and E. hyemale due to the scarce annotation from species E. xferrissii. A total of 108 training images were used to detect the nodes of the plant species with a mini-batch gradient descent method, a batch size of 8 and a learning rate of 0.0001. Various data augmentation including rotation and random cropping of a 500x500 image were used while training the model for 2000 iterations. From the detected plant nodes statistical features involving several nodes present for each specimen were used to training a simple KNN classifier. Using a separate test set of 30 images, the author reported a classification accuracy of 90%.

Ledesma et al [59] proposed a novel architecture to automatically detect color reference charts (CRC) from the digitized herbarium specimen image. As opposed to existing state-of-the-art object detection models such as R-CNN or Faster R-CNN, the study proposed a domain-specific neural network that improved the detection performance of smaller CRC together with the reduced processing time. The proposed pipeline used a sliding window together with a simple square searching algorithm to locate the possible region of the CRC. From the detect regions, the color histogram was computed for both RGB and HSV color space and the extracted features were used to train a simple MLP. This approach offers several advantages over existing CNN implementations. (1) The use of a simple sliding window was way faster than training the whole regional proposal network of the Faster R-CNN which would be much more complex. (2) The use of color histogram features eliminated the size constrains of the CRC as it produced a fixed dimension of features regardless of the input size. (3) The extracted features are then processed using a simple MLP which is less computationally expensive compared to CNNs. The study used a total of 3344 training samples and 988 test samples. Various image augmentation techniques were applied to improve the training samples including darkening, desaturation, shifting, rotation and even overlaying the CRC on different place of the same image. The study reported accuracy of 98.053% which is slightly lower than the modified Faster R-CNN (99.241%) but with a lesser processing time. Similarly, Ott et al. [61] proposed an object detect pipeline to extract leaves from herbarium images. Using an open-source labelling tool (LabelImg), a total of 889 intact leaves were annotated from 243 herbarium images. The study trained a Faster R-CNN model using an Inception-ResNet architecture as a backbone network. The study reported an accuracy of 95% on a sub-set of 61 test images.

Triki et al. [67] proposed numerous improvements over existing state-of-the-art deep learning object detection model YOLO V3. These improvements included a new added feature map scale to improve the detection of small objects and increasing the up-sampling layer(4x) to capture a higher resolution feature map. The study trained the model from a total of 4000 manually annotated herbarium images to detect 7 different categories including color charts, barcode, herbarium stamp, specimen envelope with loose material, scale bar, herbarium label together with the plant specimen. The network was trained for 10000 iterations with a batch size of 6 and a learning rate of 0.0001. Using the improved YOLO V3, the author reported an increased performance of 93.2% mAP over 90.1% from the original YOLO V3 model.

While Triki et al. used a YOLO V3 as their baseline model, Younis et al. proposed a Faster R-CNN model to detect size different plant organs [33]. The study used a total of 653 images with manual annotation of 19654 b-boxes for different plant organs. Faster R-CNN model was then trained on a subset of 498 training images using a stochastic gradient descent (SGD) optimizer with a learning rate of 0.0025 on a Detectron2 framework. Using the coco evaluation metrics, the model achieved an AP50 of 32.1 and AP75 of 16.1 for separate testing data. As reported by previous studies [62], the model worked well for large plant organ such as leaves but struggled to detect smaller organs such as seed.

### 3.4.5. Optical Character Recognition

Optical character recognition (OCR) is one of the earliest computer vision tasks tried by the researcher to solve the text extraction and digitization problem. An OCR system is a trained model that can recognize the different characters in an image [104]. The model is usually trained on a large number of sample characters to learn the structure and look for those specific characters. Once the model is well trained, a text from an image can then be detected and extracted. Popular OCR systems like Tesseract are trained using deep neural networks such as Long short-term memory (LSTM) with thousands of characters [105]. The trained model is then structured as a pipeline together with other image processing techniques such as segmentation and contour detection to detect, extract and recognize the new characters [106]. From the reviewed articles, we identified 5 studies which have investigated various aspects and uses of OCR system during digitization of herbarium specimens. The earlier studies of [104] and [107] have used the ABBYY OCR system while some of the recent studies have adopted the current-state-of-the-art Tesseract OCR system.

For example, Drinkwater et al. [104] Investigated the application of the OCR system to automate and speed up the data entry process during digitization of herbarium specimen. The ABBYY OCR software was used and the investigation was carried out on 20,000 herbarium sheets. The study found out that the use of OCR was more useful for labels with long descriptions. However, the difference in time compared to the manual was small when the label description was short. The authors further suggested that insignificance difference in time was attributed to the output produced by the OCR system which could not allow users to copy multiple lines of detected text.

To handle OCR errors, Heidorn and Zhang [108] proposed to improve the structure of the OCR outputs by performing preprocessing steps using a fuzzy-match algorithm. The author used a Hidden Markov Model (HMM) to deal with the output errors from the OCR system. These errors together were replaced with constants tokens and the proposed tool (LABELX) generated a structured XML data and RDF. Similarly, Barber et al. [107] [109] proposed a semi-automated digitization workflow that incorporated the use of OCR system. Specifically, the authors used the ABBYY FineReader OCR system along with other software such as word processing, text parser and barcode renaming tool to streamline the digitization workflow. Other works such as [41] have proposed the use of segmentation models to locate possible text from the digitized herbarium specimen to improve the OCR performance. The authors used the Tesseract version 4 and the results suggest an improved accuracy and speed of the OCR system when presented with only a portion of the image containing text rather than the whole herbarium image. Despite numerous advancements in the OCR system, the adaptation of existing OCR system to completely automate the data capturing or data entry from digitized herbarium specimens is relatively low [110]. Most of the existing studies have relied on manual entry or public participation although these could be more costly as the number of specimen increases [109].

### 3.4.6. Generative Models

Generative adversarial networks (GANs) are families of deep learning model designed by jointly training two networks where one of the networks is trained to generate the desired output (Generator network) while the other network (Discriminator network) is used to evaluate the generated output from the generator to check whether the generated output is real (coming from the ground truth) or fake (generated by the generator network) (Fig 9) [111]. GANs have shown numerous successes in various domains and applications and these can be trained to perform various computer vision tasks such as image-to-image translation, segmentation or generating synthetic images [112], [113]. In this review, we identified only 3 studies that have attempted to use GANs for different tasks related to image-to-image translation and domain adaption [37], [38], [53].

To improve the cross-domain compatibility between herbarium specimen and their field samples, Juan et al. [38] proposed the use of GAN networks to perform reverse senescence of herbarium specimen. In their study, three different species each with 20 samples of their fresh leaf and another 20 samples of their senescence leaf were used to train a GAN network to perform image-to-image translation from a senescence to fresh leaves [114]. The study used 80% of the data to train a Pix2Pix model and reported a 0.9 structure similarity index measure (SSIM) achieved when using the remaining 20% for testing. Similarly, Hussein et al. [37] proposed deep learning approaches to reconstruct damaged herbarium leaves. In their study, the authors compared the performance of Pix2Pix GAN network with a more recent state-of-the-art Pconv model for image inpainting. Both Pix2Pix and PConv model achieved a SSIM of 0.9 at different damage threshold. The authors demonstrated the feasibility of improving the identification accuracy when using their proposed approach to reconstruct the damaged leaves.

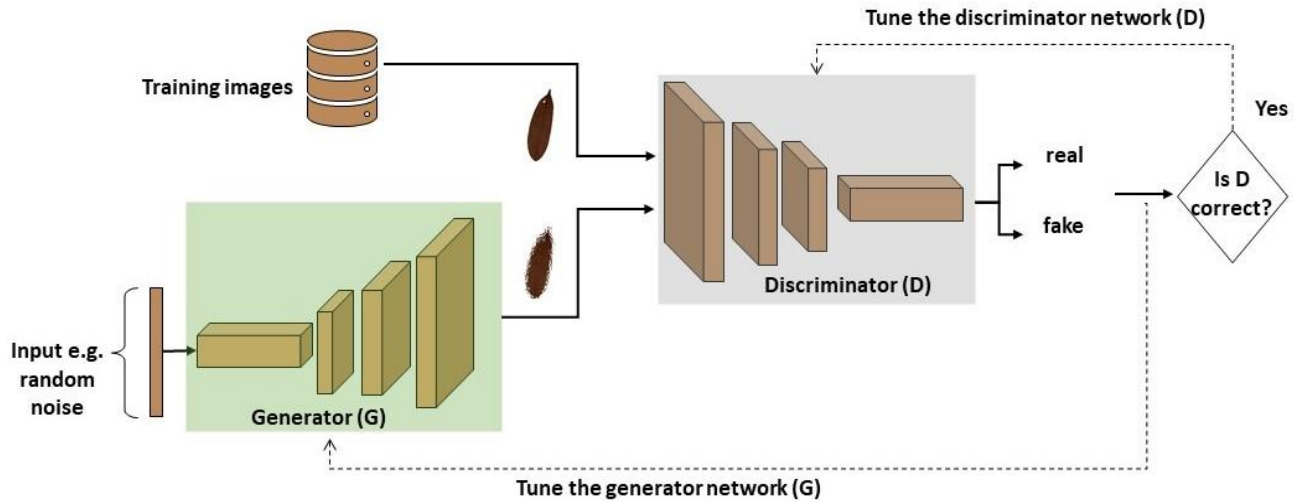

Figure 9: Architecture of GAN network

To reduce the domain gap between fresh and dried plants, Villacis et al. [53] proposed the use of GAN network based on a few-shot adversarial domain adaptation to perform cross-domain adaptation between fresh and herbarium plants [115]. Using their proposed approach, the authors emerged as the winners of a challenging cross-domain species identification from the PlantCLEF2020 competition [56].

### 3.4.7. Digitization Workflows

Digitization of fragile specimens, such as those of herbarium, is both an expensive and a time-consuming task. A typical digitization workflow involves several steps including image capturing, information extraction, storage, etc. Different standards and guideline have been proposed to ensure the usability of the produced images on a large pool of application with possible unlimited life-span [116]. We identified 5 studies which have specifically focused on the use of CV and ML techniques to automate and streamline various stages of digitization workflow. These studies were categorized as workflows as they specifically focused on using existing implementations/models to improve the overall digitization pipeline.

Hidalga et al. [116] proposed a new digitization workflow that focuses on the quality of the images acquired and stored. Authors incorporated their previous proposed study of using semantic segmentation to speed up the audits of quality control and information extraction [70]. Studies such as [73] have attempted to combine various CV and ML techniques to automate the data extraction process from digitized herbarium specimen. The authors proposed the use of existing trained segmentation model to localize and identify various objects from the images, identification CNN model to detect the presence of certain plant organs, an OCR technique to capture related labels and a species recognition model to identify possible misidentification and unidentified species.

### 3.5. Prototypical Implementation (RQ-5)

The prototypical implementations of CV and ML techniques for digitized herbarium specimens have mostly focused on automating the digitization workflows and the analysis of digitized specimens. As shown in Table 6, there have been 11 studies which managed to implement the proposed approaches as a software package. Out of these studies, 3 studies have been solely published as a software tool. These packages have mainly implemented as a desktop application due to the massive image size of the specimens. Among these, segmentation approaches have been the most utilized techniques as a preprocessing step to extract various types of information from herbarium specimen [69], [31], [117], [118], [68]. OCR technique has been mainly used to automate some of the tasks during the digitization workflow such as detecting specimen labels and language transcription [119], [120], [35]. Although object detection is considered a less complex task compared to pixel-wise segmentation, only a single study has been found which their prototype has been developed based on object detection technique [35].

As part of the software development package named MORPHIDAS[4], Corney et al. [69] initially proposed an algorithm to automatically detect leaves from herbarium specimen images. The algorithm uses a deformable template approach optimized by evolutionary algorithms to segment the leaves. Morphological operations are then applied to detect the main vein and extract the length and width of the leaf. As part of incremental improvement of the overall package, Corney et al. [31] proposed an algorithm to automatically locate the leaf teeth on leaf margin using various image processing techniques. Once the teeth of the leaves have been located, the algorithm can calculate different teeth features such as area, perimeter, teeth angles and counting the teeth. The two algorithms were implemented and tested as software using Matlab version 7.10.0 (R2010a) with Image Processing Toolbox.

Barber et al. [107] proposed a semi-automated label information extraction workflow (SALIX method) to automate capturing of labels information from specimen images. The proposed software incorporates different packages such as ABBYY FineReader an OCR system, word processing packages and Adobe Lightroom for image management with the windows-based executable program interface. The software is linked to Symbiota data portal (SEINet) for providing web access to data and images [107].

Sweeney et al. [120] proposed an automated conveyor system for the digitization of herbarium specimen collection. The software integrates different components during digitization workflow to capture the specimen image and the data associated with the specimen image. Kirchhoff et al. [35] proposed automated digitization workflow to capture and extract information from the herbarium specimen image. The proposed software is built on top of OpenRefine an open-sourced java-based software tool for cleaning and processing data. The core computer vision techniques driving the software are a template matching algorithm [121] that is used to detect the scale in an image, a line contrast approach that takes advantage of alternating contrast in the text to detect text regions and an OCR system (Tesseract/OmniPage) that is used to convert detected text regions of the images into texts. The extracted information is further processed by different text-based algorithms to capture and store required data. The developed software is provided as a web service using a framework called swagger[5].

To improve the availability of annotated herbarium specimen data, Bouaziz et al. [117] developed an open-source java-based desktop software tool for annotation of digitized herbarium specimens named as Specimen-GT tool. The tool enables users to generate bounding boxes for a different part of the specimen. Using an active snake algorithm, the tool can automatically assist in detecting the leaf shape which can be used to extract various plant traits. The tool can be used both offline with a local repository or integrated with an online repository in which the annotation or extracted traits can then be exported in various formats such as XML and CSV. Another similar software is the TraitEx which was developed by [118]. TraitEx is an open-source standalone java application that is used to extract various leaf shape features such as length, width and area. This software is further integrated with another image processing software named imageJ to enable further preprocessing and to edit the specimen image. Similar to [117], the extracted features can then be exported in a CSV format for further processing. Studies such as [122] have started to use the tool for assessing trait variability on herbarium specimens.

---

[4] http://www.computing.surrey.ac.uk/morphidas/
[5] https://swagger.io/

Table 6: Prototypes Implemented based on proposed approaches. *integration with herbarium specimen collection

| Name | Application | Task | Techniques | Availability | Computation | year | Reference |
|---|---|---|---|---|---|---|---|
| MORPHIDAS: ToothFinder | Desktop | Analysis | Segmentation | https://github.com/dcorney/ToothFinder | Offline | 2012 | [31] |
| MORPHIDAS: LeafFinder | Desktop | Analysis | segmentation | http://www.computing.surrey.ac.uk/morphidas/leafFinder.html | Offline | 2012 | [69] |
| SALIX Method | Desktop | workflow | OCR | - | - | 2013 | [119] |
| - | Desktop | Workflow | OCR | https://github.com/psweeney-YU/NEVP-conveyor | Offline | 2018 | [120] |
| StanDAP-Herb | Web-service | Workflow | OCR, template matching, text detection algorithm | http://api.bgbm.org/standap/download/openrefine-extension | - | 2018 | [35] |
| Specimen-GT tool | Desktop | Analysis | Object detection, active snake segmentation | - | Online/offline | 2018 | [117] |
| TraitEx Software | Desktop | Analysis | Segmentation | https://bitbucket.org/traitExTool/traitextool | Offline | 2019 | [118] |
| GinJinn | Desktop | Analysis | Object detection | https://github.com/AGOberprieler/ginjinn | Offline | 2020 | [35] |
| LeafMachine | Desktop | Analysis | Segmentation | https://github.com/Gene-Weaver/LeafMachine | offline | 2020 | [68] |
| SYNTHESYS+ | Desktop/web-service | Workflow | OCR, segmentation | - | Online/offline | 2020 | [123] |
| HerbASAP | - | Workflow | - | https://github.com/CapPow/HerbASAP | - | 2020 | [59] |

GinJinn is another software tool developed by Ott et al. [61] to detect and extract various part of herbarium specimen such leaves and flowers. The core computer vision technique is the object detection model based on CNN. The software was developed using python language with Tensorflow framework. The proposed software enables users to supply and train the detection model or use an existing detection model. The software is open-sourced and can be accessed as an API. A similar software LeafMachine offers the same functionality but using a more precise approach to detect and extract features from a single intact leaf [68]. The algorithm uses a trained DeepLabv3+ model to perform a pixel-wise segmentation of specimen leaves from the rest of the background and then uses a trained SVM model to select whether the detected leaf was an individual leaf or not. The software was developed and tested using Matlab R2019b which can run on various operating systems.

Most of the existing software systems are desktop applications due to the nature of the required functionality. A mobile application that can run on hand-held or other portable devices could further facilitate better usage of the specimen images for the task such as species identification or other similar studies. Open-sourcing current and future software for analysis of herbarium specimen could enhance/ enrich current biodiversity databases.

Walton et al. [123] proposed SYNTHESYS+ which is a digitization workflow platform initiative that was designed as a tool for the digitization of natural history collections. The proposed platform aggregates submodules ranging from image segmentation, object detection, OCR system, specimen trait extraction and identification. Preliminary gap analysis revealed that some of the key modules, such as trait extraction and specimen feature analysis techniques, have received less attention compared to other modules such as segmentation, object detection and OCR systems. The platform is currently in the preliminary phase of its implementation.

Ledesma et al. [59] highlighted a program named HerbASAP as a possible platform for deploying their model. HerbASAP is a large digitization program that is currently under development although no much details were found regarding the current status of the project.

### 3.6. Challenges and Opportunities (RQ-6)

In this section, we highlight some of the existing challenges within the reviewed literature and offer some alternative solutions. These challenges encompass both on the dataset level and at the technique level.

### 3.6.1. Specimen Data Deficient (Unbalanced)

Tropical regions are characterized by having the most diversity of plant species [124]. Digitization of these specimens has enabled a massive availability of image dataset to train deep learning models. Despite the benefit of large available samples, some of the important plant specimens are still represented by a very small amount of images which affect the overall performance of the models [5]. Furthermore, other factors such as sampling biases, digitization and incorrect identification of species may have further increased the uneven distribution of data [13], [19], [78]. Techniques such as data augmentation have been used as a possible way to improve model generalization despite availability of small samples for some species [55]. These techniques include random cropping of specimen images, changing image contrast, brightness, color jittering, flipping, rotation or more recent domain-specific augmentation such as centre tilting with random zooming of herbarium images [53], [125]. Transfer learning is another technique that has been commonly used in the literature. This technique involves initialization of deep learning models with weights of trained models from other domains and has been demonstrated more beneficial than initializing with random weights [5],[55]. New approaches to this problem are the use of field images along with herbarium image to train deep learning models (cross-domain identification) [56], but it remains an open research area that needs further investigation.

### 3.6.2. Cross-Domain Species Identification

PlantCLEF2020 challenge has demonstrated the difficulties of combining untapped resources of herbarium images with their field counterparts in training deep learning models [56]. Nevertheless, new techniques such as domain adaption [53] and Siamese Convolutional Networks (SCN) [52] have demonstrated their superior performance over traditional CNN networks. It will be of great interest to see how these approaches evolve overtime while exploring new architectural designs such the one proposed by [126] to improve the performance. Although it's a proof of concept study [38], the use of GANs to perform reverse senescence of herbarium specimen will be of great interest in the near future.

### 3.6.3. Architecture Dedicated for Species Identification

Ensemble learning has been commonly adapted when involving a large number of species to identify [55]. Despite the improvements of ensemble learning over the single model, these models are computationally intensive to train and require a trial-and-error approach to get a balance between the number of ensembles versus performance tradeoff. As suggested by Carranza et al. [5], architecture-specific deep learning models dedicated to species identification needs to be further investigated. Furthermore, the use of taxonomic knowledge during the training of deep learning models by either hierarchical learning or using dedicated loss functions is another area that needs further research as only a few studies have attempted to investigate it [50], [127].

### 3.6.4. Computational Complexity of Training CNNs

Unlike many domains, herbarium images are usually in high resolution. Many existing CNN implementations have scalability issues when training with high-resolution images due to the increase of computational complexity. New approaches that incorporate image and network statistics could reduce the computational cost while using higher resolution images [51]. Techniques such as training models in the frequency domain instead of the spatial domain could also improve the performance and enable training of deeper models with higher resolutions [128]. Even though ensemble learning has shown promising performance in herbarium species identification [55], techniques such as knowledge distillation or network pruning should further be investigated to possibly reduce the size of these model and make them more practical in real-world settings [129].

### 3.6.5. Segmentation of Plant Organs is still a Challenging Task

Despite its importance in phenological studies [3], segmentation of specimen organs such as seeds, fruits, flowers and leaves is still a challenging task [62], [63]. Due to the pressing and drying process, these organs have lose their natural structures and appearance and hence present a challenge even for the current-state-of-the-art deep learning models [84]. A larger training dataset with a diverse number of taxa involved could improve the accuracy of the models [68], [63]. On the other hand, organ-specific models need to be further investigated as they have shown improvements over jointly training a single model for segmenting multiple organs [68].

### 3.6.6. Ground Truth Annotation for Training Segmentation Models

Generating ground truth annotation for segmentation task is both labour-intense and a time-consuming task. Most of the existing studies on segmentation tasks have used a relatively smaller training sample for a limited set of taxa [65]. Techniques such as active learning proposed by Mora-Fallas et al. [34], few-shot learning [130], [131], semi-supervised learning [132], [133] and weakly supervised learning [134] may provide a new approach in working with a small number of training samples. On the other hand, techniques such as domain randomization could also be investigated in producing synthetic herbarium images together with its ground truth annotation as this technique has demonstrated its applicability in other domain areas including [135] and [136]. Data augmentation techniques have also shown promising results when used for training object detection models [59]. Moreover online participation through citizen science projects to enhance and speed up the annotation process [109], [137].

### 3.6.7. Interpretability of Deep Learning Models

Despite the remarkable performance of deep learning techniques for herbarium species identification, they have always been advocated as black-box since it's hard to interpret the decision of the models [138]. On the other hand, hand-crafted features are commonly used as they incorporate domain-specific knowledge although their scalability is limited [7]. None of the existing studies that has attempted to investigate/interpret the decision made by the deep learning models. The insights offered by these models can enable not only computer scientists but also taxonomist/botanist to discover perhaps novel and previously unknown plant features. Furthermore, new insights can further help in designing hybrid features which can improve the performance of the models [139].

### 3.6.8. Micro-level herbarium species identification

Even though most of existing digitized herbarium collections consist of high-resolution images, deep learning techniques have mostly been used on downscaled images which could potentially lead to loss of some important botanical features. For a botanist to successfully identify a specimen, a microscopic view of the specimen is needed to reveal some of the subtle features of these specimens [7], [140], [141]. This is especially important for the existing dried collections as most of these collections have lost their visual characteristics due to the pressing and drying process [75]. This could be an important step in herbarium species identification if the burden of collecting microscopic images of these specimens is overcome [142]. Furthermore, using micro-level images could also help in overcoming other challenges such as species sample deficiency, cross-domain identification or even using higher taxonomic level features such genus and family to identify new species in particular taxa [143].

### 3.6.9. New Collection and storage procedures for herbarium specimens

Despite the revolution of CV and ML techniques for digitized herbarium specimens, the procedure for collection and preservation of these valuable specimens has not changed for a while [140]. Existing collection and preservation procedures of these specimen were only intended for human physical inspection [144]. New approaches that could enhance and maximize the use of these valuable collections are needed. Collection of these specimens should foster the next-generation of data-driven research by including plant meta-data, diverse specimen types, different specimen organs and different views of the specimen while capturing other ancillary data such as phenotypic or genotypic information whenever possible [137], [145], [146], [147], [148]. This could also improve the usability of the specimens for different tasks such as identification process rather than relying only on a single image of the herbarium sheet. None of the reviewed studies has attempted to use any extra meta-data alongside images to build an identification system.

### 3.6.10. Quality Herbarium images for Computer vision and machine learning research

Not all herbarium specimen images can be used for training machine learning models. Some of these specimens consist of only a portion of the plant which may not present an important feature for training the model. On the other hand, due to the drying and pressing process, some of these specimens can be damaged and lose their morphological structure that can be necessary for training models. For other species, parts of plants such as reproductive structures can only be exposed by dissecting the plant part which can limit their usability for training phenological scoring models [84]. Apart from the natural issues of these plants, presence of other non-plant objects such as color charts, scale bar, barcodes and plant labels may induce noise/biases in training machine learning models. This is likely because the preparation and digitization of these specimens varies in different herbaria hence the models may learn from the noise and not from the plants itself [19]. Finally, as stated in [13], misidentified specimens or duplicate specimen images is another thing that researchers should be aware of. It is therefore recommended to ensure the quality and quantity of the training data are maintained by collecting a diverse set of specimen images.

### 3.6.11. Global Leaf Dataset from Herbarium Specimen Collections

For species identification tasks, a leaf is considered as an important botanical feature for differentiating among species as it is mostly present throughout the season [149],[150],[151]. A leaf carries many unique features of the plant including the color, shape and the texture which vary among species and hence make it the widely used plant organ for identification task [26]. Most of the existing studies have been performed on the benchmarked leaf dataset of fresh plants collected and contributed by various researchers [152]. Herbarium plant species identification has received less attention from CV community as collecting samples of individual leaves is not a trivial task due to existence of noise and their sheer size with millions of digitized specimens already stored in different repositories. Since there already exist a large number of digitized herbarium images, therefore it is necessary to further invest on different techniques which can be used for extraction of individual leaves. This will have multiple benefits for both scientific research as well as botanical research for different studies such as plant ecology and physiology [118]. Some attempt have already been made to extract these leaves although the existence of overlapping and damaged leaves needs to be further investigated [68].

### 3.6.12. Benchmarking Dataset for Object Detection and Segmentation for Herbarium Specimens

Unlike other domains where there already exist some benchmarking dataset for detection and segmentation tasks (example MSC COCO, Cityscapes, ADE20K and Pascal VOC datasets [153]), herbarium specimen images lack similar benchmarking dataset which may prevent fast progression of the field. This is likely caused by few factors: (1) Herbarium specimens vary significantly hence selecting appropriate specimen dataset that could generalize the representation of specimens in the different herbarium is challenging and (2) the tedious annotation process of different plant organs that are present in herbarium specimens. Effective collaboration needs to be initiated to ensure important benchmarking dataset are available rather than working on individual datasets.

### 3.6.13. Standardized Metrics for Reporting Results

Evaluation of proposed methods is an important step for assessing improvements over existing studies. From the reviewed literature, it was found that several metrics are being commonly used. For example for species identification task, both accuracy and a Mean Reciprocal Rank (MRR) used in various studies [5],[56]. Other studies involving segmentation tasks have only used a single metric (AP.50) which may introduce bias in the interpretation of results [62]. It is therefore recommended for studies to report multiple metrics such as AP at different threshold ($AP_{small}$, $AP_{medium}$, $AP_{large}$, AP.50, AP.75, AP.90) and provided a breakdown for different classes involved such as in [62] or reporting accuracy at different rank if the study involves a large number of species. Further investigation needs to be done to assess the impact of evaluation metrics used with other constraints such as an imbalanced class problem or a varying number of test samples per species problem [154].

## 4. Discussions

In this study, we have provided a comprehensive systematic review of the applications of CV and ML for digitized herbarium specimens and series of research questions were formulated and a systematic procedure was then followed to answer these questions. We present here the summary of the main findings of this study and offer possible future direction whenever possible.

## 4.1. Findings-1: Difficult of Comparing and Evaluating Proposed Methods

Wäldchen et al. [26] have already reported on the difficulty of comparing and evaluating proposed methods for plant identification task. We also found the same difficulties in comparing different proposed methodologies. This is due to several reasons (1) Almost all studies have used different datasets to evaluate their proposed methods (2) Even when the datasets used are made publicly available, these datasets are either missing their ground truth annotation or the authors have only public a part of the dataset (3) There do not exist any benchmarking datasets to test the new methodologies. The only complete datasets available are for species identification task which are provided by [55], [56]. Different studies have reported different metrics which makes the comparison more challenging. As stated in the previous sections, a need for the standardized metric is required to ensure incremental progress is made by new proposed methodologies.

## 4.2. Findings-2: Most of the studies have Used Deep Learning

We also found out that deep learning techniques are increasingly being used compared handcrafted features. This is expected due to the current increase in digitization effort and the amount of public available herbarium data. Furthermore, challenging tasks such as phenological feature extraction which could not be successfully performed using traditional image processing techniques are now possible when using deep learning approaches.

## 4.3. Findings-3: Transfer Learning has been the common approach

Out of all studies which have used deep learning, only one study has proposed a custom CNN architecture and trained the network from scratch [59]. Pre-trained models including ResNets, VGG and GoogleNet have been the most popular choice of models for different tasks. On the other hand, DeepLabv3+ and Mask R-CNN have been used in all studies related to the segmentation task except one study [64]. For object detection task, different models have been applied in the literature. This clearly shows the lack of superiority of one model over the others. Despite the robustness of these models, currently no model which has proved to be more robust in term of both accuracy and efficiency for other domain. Therefore, there is a need to further investigate new techniques which could prove more useful in the near future.

## 4.4. Findings-4: Identification Studies have Achieved more Success Compared to Phenological studies

Among all the computer vision task investigated, species identification has achieved the most successful performance compared to other tasks such as object detection or segmentation. This is likely because of the huge amount of data available and the ease of curation for identification task compared to preparing of ground truth annotation for object detection and segmentation task. Extraction of phenological traits is challenging due to the nature of herbarium specimens. The accuracy of detecting phenological features can be further improved by working with higher resolution images as these features are sometimes hard to see even with a human physical inspection.

## 4.5. Findings-5: Biology Related Journals and Conferences Have Been the Main Platform for Publication

With exception of [7], [46], [51], [38], [67], [65] and [71] all other studies have been published in conferences or journals related to plant science. Among the popular journals includes Taxon, Application in plant science, The Biodiversity Data Journal, and the Biodiversity Information Science and standards with all been published as an open-access publication. There has also been a trend of publishing an early preview of the work through extended abstracts publication. In this study, we have reviewed 6 extended abstracts which were all published using the Biodiversity Information Science and Standards journal.

## 4.6. Findings-6: There exists a strong collaboration between Computer Scientists and Biologists

Due to the steady progress of machine learning and computer vision techniques, both computer scientists and biologists are working together to solve novel problems which were not seen before [148]. From the reviewed literature, the majority of studies involved more than one academic discipline which may have helped for the current rapid progress in the field. It is therefore encouraged to increase an inter-disciplinary approach to ensure effective collaboration in solving important challenges.

## 5. Conclusion

In this study, the existing literature on applications of computer vision and machine learning techniques has been reviewed and discussed. While herbarium serves as a crucial data source for biologist, it's also presents a great opportunity for computer scientist due to increasingly availability of their digitized image data. We argue that, with the current open challenges discussed in this study, more effort is needed to fully realize the potential of this invaluable data. CV and ML techniques are currently the best tools that need to be utilized to speed up the automation and uncovering of hidden patterns that exist in these century old data. This will have a direct implication to our current environmental sustainability goals hence more interdisciplinary research between computer scientist and biologist is need.

**Acknowledgements**

This work is supported by Universiti Brunei Darussalam under research grant number [UBD/RSCH/1.4/FICBF(b)/2018/011].

**Declaration of Interest**

All authors declared that there is no any form of financial or personal interest that may have influence this work.